\documentclass[11pt,a4paper]{article}
\usepackage[hyperref]{acl2018}
\usepackage{times}
\usepackage{latexsym}

\usepackage{url}

\usepackage{graphicx}
\usepackage{amsmath}
\usepackage{amsfonts}
\usepackage{graphicx}
\usepackage{verbatim}
\usepackage{booktabs}
\usepackage{subfig}
\usepackage{amsthm}
\usepackage{amssymb}
\usepackage{qtree}
\usepackage{graphicx}
\usepackage{caption}
\usepackage{multirow}
\usepackage{mathrsfs}
\usepackage{tablefootnote}
\usepackage{arydshln}
\usepackage{lscape}
\usepackage{afterpage}
\usepackage{todonotes}
\usepackage{arydshln}
\usepackage{color}
\usepackage[linewidth=1pt]{mdframed}

\DeclareMathOperator*{\argmax}{arg\,max}

\usepackage{url}
\usepackage{caption}
\usepackage{xcolor}
\usepackage{calc}
\def\mybar #1#2#3{
  {\color{red}\rule{#1cm}{4pt}} & {\color{green}\rule{#2cm}{4pt}} & {\color{blue}\rule{#3cm}{4pt}} }

\usepackage{pgfplots} 
  
\aclfinalcopy 

\newcommand\numberthis{\addtocounter{equation}{1}\tag{\theequation}}

\title{Improving Entity Linking by \\ Modeling Latent Relations between Mentions}
\author{Phong Le$^1$ and Ivan Titov$^{1,2}$ \\
  $^{1}$University of Edinburgh   $\;\;^{2}$University of Amsterdam \\
  {\tt \{ple,ititov\}@inf.ed.ac.uk}
  \\}

\date{}

\begin{document}
\maketitle
\begin{abstract}

Entity linking involves aligning textual mentions of named entities to their corresponding entries in a knowledge base. Entity linking systems often exploit relations between textual mentions in a document (e.g., coreference) to decide if the linking decisions are compatible. Unlike previous approaches, which relied on supervised systems or heuristics to predict these relations, we treat relations as latent variables in our neural entity-linking model. We induce the relations without any supervision while optimizing the entity-linking system in an end-to-end fashion. Our multi-relational model achieves the best reported scores on the standard benchmark (AIDA-CoNLL) and substantially outperforms its relation-agnostic version. Its training also converges much faster, suggesting that the injected structural bias helps to explain regularities in the training data.

\end{abstract}

\section{Introduction}
\label{sec:intro}

\begin{figure*}[ht]
\centering
\includegraphics[width=1.\textwidth]{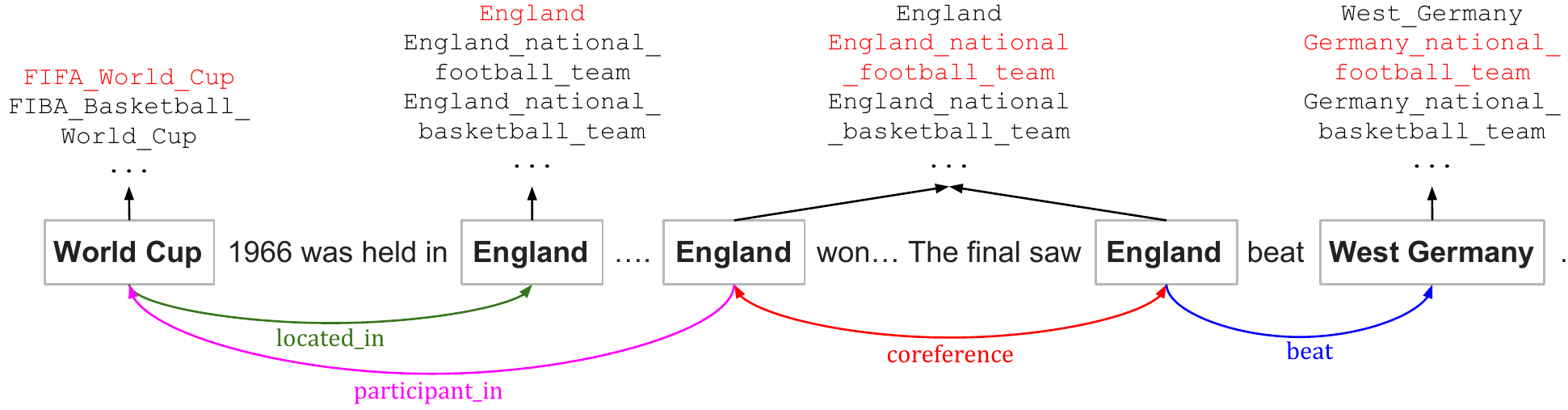}
\caption{Example for NEL, linking each mention to an entity in a KB
(e.g. ``World Cup'' to \textsc{FIFA\_World\_Cup} rather than 
\textsc{FIBA\_Basketball\_World\_Cup}). 
Note that the first and the second ``England'' are in different relations to ``World Cup''.
}
\label{fig:multi-rel example}
\end{figure*}

Named entity linking (NEL)
is the task of assigning entity mentions in a text to corresponding entries in a knowledge base (KB). For example, consider Figure~\ref{fig:multi-rel example} where a mention ``World Cup'' refers to a KB  entity \textsc{FIFA\_World\_Cup}.
NEL is often regarded as crucial for natural 
language understanding and commonly used as preprocessing for tasks such as information extraction \cite{hoffmann-EtAl:2011:ACL-HLT2011} and question answering \cite{yih-EtAl:2015:ACL-IJCNLP}.

Potential assignments of mentions to entities are regulated by semantic and discourse constraints. For example, the second and third occurrences of mention ``England'' in  Figure~\ref{fig:multi-rel example} are coreferent and thus should be assigned to the same entity. Besides coreference, there are many other relations between entities which constrain or favor certain alignment configurations. For example, consider relation participant\_in in Figure~\ref{fig:multi-rel example}: if ``World Cup'' is aligned to the entity \textsc{FIFA\_World\_Cup} then we expect the second ``England'' to refer to a football team rather than a basketball one. 

NEL methods typically consider only coreference, relying either on off-the-shelf systems or some simple heuristics \cite{TACL637}, and exploit them in a pipeline fashion,  though some (e.g., \newcite{cheng-roth:2013:EMNLP, ren2017cotype}) additionally exploit a range of syntactic-semantic relations such as apposition and possessives. 
Another line of work ignores relations altogether and models  the predicted sequence of KB entities as a bag~\cite{P16-1059,K16-1025,D17-1276}.
Though they are able to capture some degree of coherence (e.g., preference towards entities from the same general domain) and are generally empirically successful, the underlying assumption is too coarse. For example, they would favor assigning all the occurrences of ``England'' in Figure~\ref{fig:multi-rel example} to the same entity.

We hypothesize that relations useful for NEL can be induced  without (or only with little) domain expertise. In order to prove this,
 we encode relations as latent variables and induce them by optimizing the entity-linking model in an end-to-end fashion. In this way, relations between mentions in documents will be induced in such a way as to be beneficial for NEL. 
As with other recent approaches to NEL \cite{yamada2017learning,D17-1276}, we rely on representation learning and learn embeddings of mentions, contexts and relations. This further reduces the amount of human expertise required to construct the system and, in principle, may make it more portable across languages and domains. 

Our multi-relational 
neural model achieves an improvement of 0.85\% F1 over the best reported scores on the standard AIDA-CoNLL dataset~\cite{D17-1276}.
Substantial improvements over the relation-agnostic version
show that the induced relations are indeed beneficial for NEL.
Surprisingly its training also converges much faster: training of the full model requires ten times shorter wall-clock time than what is needed for estimating the simpler relation-agnostic version. This may suggest that the injected structural bias helps to explain regularities in the training data, making the optimization task easier. We qualitatively examine induced relations. Though  we do not observe direct counter-parts of linguistic relations, we, for example, see that some of the induced relations are closely related to coreference whereas others encode forms of semantic relatedness between the mentions.

\section{Background and Related work}

\subsection{Named entity linking problem}
Formally, given a document $D$ containing a list of mentions $m_1, ..., m_n$,
an entity linker assigns to each $m_i$ an KB entity $e_i$
or predicts that there is no corresponding entry in the KB (i.e., $e_i=\text{NILL}$).  

Because a KB can be very large, 
it is standard to use an heuristic to 
choose potential candidates, eliminating options which are highly unlikely.
This preprocessing step is called 
\emph{candidate selection}. The task of a statistical model is thus reduced to choosing the best option among 
 a smaller list of candidates 
$C_i = (e_{i1}, ..., e_{il_i})$.
In what follows, we will discuss two classes of approaches tackling this 
problem: local and global modeling. 

\subsection{Local and global models}
\emph{Local} models rely only on local contexts of mentions
and completely ignore interdependencies between the linking decisions in the document (these interdependencies are usually referred to as {\it coherence}). 
Let $c_i$ be a local context 
of mention $m_i$ and $\Psi(e_i, c_i)$ be a local score function. 
A local model then tackles the problem by searching for 
\begin{equation}
	\label{equ:local model}
    e^*_i = \argmax_{e_i \in C_i} \Psi(e_i, c_i)
\end{equation}
for each $i \in \{1,...,n\}$
\cite{E06-1002,TACL637,yamada2017learning}.

A \emph{global} model, besides using local context  
within $\Psi(e_i, c_i)$, takes into account entity coherency. It is captured by a coherence score function $\Phi(E,D)$:
\begin{equation*}
	E^* = \argmax_{\substack{E \in C_1 \times ... \times C_n}} 
    				 \sum_{i=1}^n \Psi(e_i, c_i) + 
                     \Phi(E, D)
\end{equation*}
where $E=(e_1, ..., e_n)$.
The coherence score function, in the simplest form, is a sum over all 
pairwise scores $\Phi(e_i, e_j, D)$ \cite{P11-1138, huang2015leveraging, Q15-1011, ganea2016probabilistic,guorobust, P16-1059, K16-1025}, resulting in:
\begin{align*}
	\label{equ:global model}
	E^* = \argmax_{\substack{E \in C_1 \times ... \times C_n}} 
    				 & \sum_{i=1}^n \Psi(e_i, c_i) + \\
                      & \sum_{i \neq j} \Phi(e_i, e_j, D) \numberthis 
\end{align*}

A disadvantage of global models is that exact decoding (Equation~\ref{equ:global model}) is NP-hard \cite{wainwright2008graphical}. 
\newcite{D17-1276} overcome this using loopy belief propagation (LBP), 
an approximate inference method based on message passing \cite{murphy1999loopy}. 
\newcite{P16-1059} propose a \emph{star model}
which approximates the decoding problem in Equation~\ref{equ:global model} by approximately decomposing it into $n$ 
decoding problems, one per each $e_i$.

\subsection{Related work}
Our work focuses on modeling pairwise score functions $\Phi$ and 
is related to previous approaches in the two following aspects.

\subsubsection*{Relations between mentions}
A relation widely used by NEL systems is \emph{coreference}:
two mentions are coreferent if they refer to 
the same entity. Though, as we discussed in Section~\ref{sec:intro}, other linguistic relations constrain entity assignments, only a few approaches
(e.g., \newcite{cheng-roth:2013:EMNLP, ren2017cotype}), 
exploit any relations other than coreference.
We believe that the reason for this is that predicting and selecting 
relevant (often semantic) relations is in itself a challenging problem. 

In \newcite{cheng-roth:2013:EMNLP}, relations between mentions are extracted 
using a labor-intensive approach, requiring a set of hand-crafted rules
and a KB containing relations between entities. 
This approach is difficult to generalize to languages and domains which do not have such 
KBs or the settings where no experts are available to design the rules. 
We, in contrast, focus on automating the process using representation 
learning. 

Most of these methods relied on relations predicted by external tools, 
usually a coreference system. One notable exception is \newcite{durrett2014joint}: 
they use a joint model of entity linking and coreference resolution. 
Nevertheless their coreference component is still supervised, 
whereas our relations are latent even at training time. 

\subsubsection*{Representation learning}
How can we define local score functions 
$\Psi$ and pairwise score functions $\Phi$?
Previous approaches employ a wide spectrum of techniques. 

At one extreme, extensive feature engineering was used  
to define useful features. For example, \newcite{P11-1138}
use  cosine similarities between 
Wikipedia titles and local contexts as a feature when computing the   
local scores. For pairwise scores they exploit
information about links between Wikipedia pages. 

At the other extreme, feature engineering is almost completely replaced 
by representation learning. These approaches rely on pretrained embeddings of words \cite{mikolov2013efficient, D14-1162} and 
entities \cite{he-EtAl:2013:Short, yamada2017learning, D17-1276}
and often do not use virtually any other hand-crafted features.
\newcite{D17-1276} showed that such an approach can yield SOTA 
accuracy on a standard benchmark (AIDA-CoNLL dataset). Their local and pairwise 
score functions are
\begin{align}
\Psi(e_i, c_i) &= \mathbf{e}_i^T \mathbf{B} f(c_i) \nonumber \\
\Phi(e_i, e_j, D) &= \frac{1}{n-1} \mathbf{e}_i^T \mathbf{R} \mathbf{e}_j 
\label{equ:GH pairwise score}
\end{align}
where $\mathbf{e}_i, \mathbf{e}_j \in \mathbb{R}^d$ are 
the embeddings of entity $e_i, e_j$,
$\mathbf{B}, \mathbf{R} \in \mathbb{R}^{d \times d}$ are diagonal matrices. The mapping $f(c_i)$ applies an attention mechanism to context words in $c_i$ to obtain a feature representations of context  ($f(c_i) \in \mathbb{R}^d)$. 

Note that the global component (the pairwise scores) is agnostic to any relations between entities or even to their ordering: it models $e_1, ..., e_n$ simply as a bag of entities. 
Our work is in line with \newcite{D17-1276} in the sense that 
feature engineering plays no role in computing local and pair-wise scores. 
Furthermore, we argue that pair-wise scores should take into 
account relations between mentions which are represented by 
\textit{relation embeddings}.

\section{Multi-relational models}

\subsection{General form}

\begin{figure}[h]
	\centering
    \includegraphics[width=0.49\textwidth]{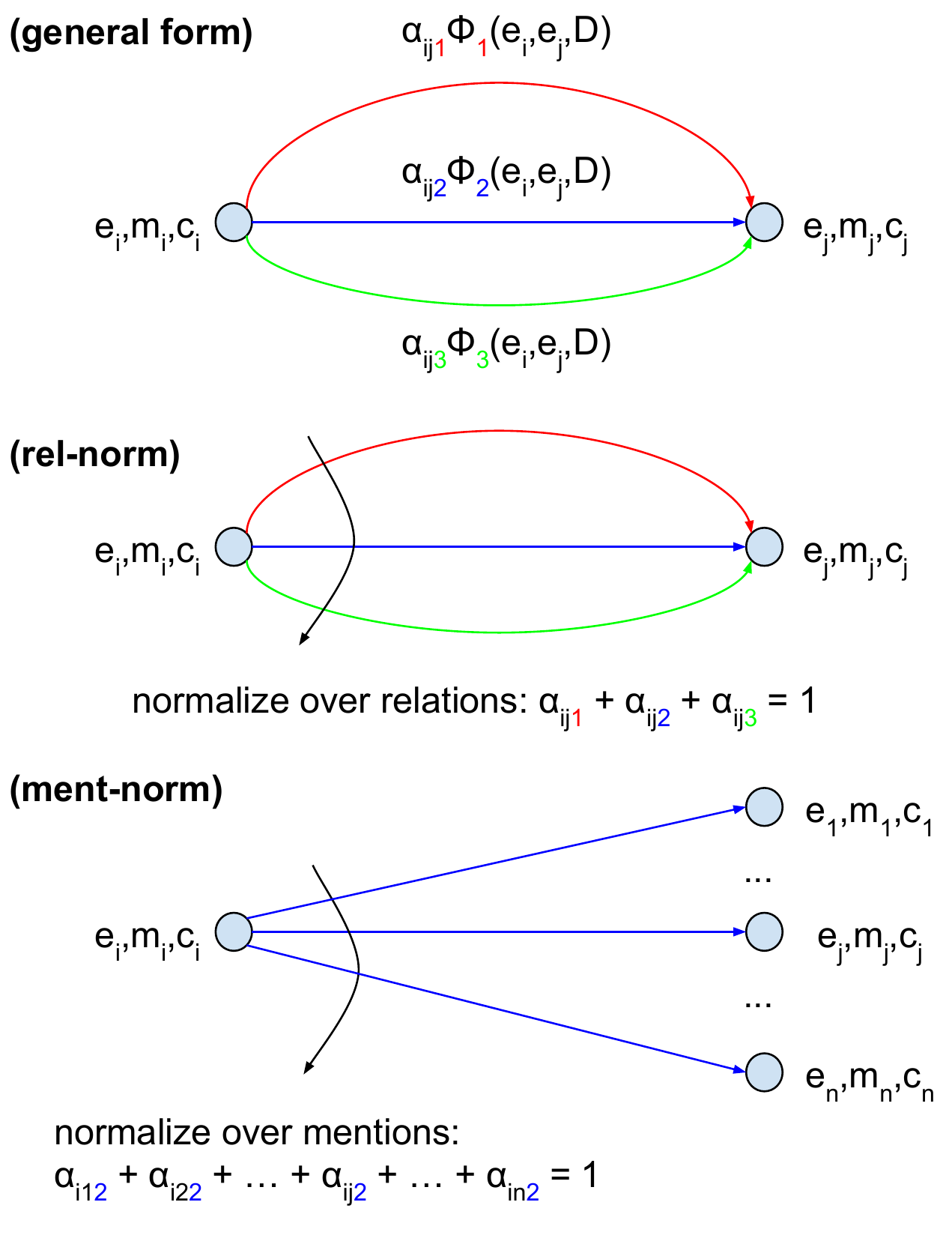}
    \caption{Multi-relational models:  general form (top), 
     rel-norm (middle) and ment-norm (bottom). 
    Each color corresponds to one relation.}
    \label{fig:multi-relation}
\end{figure}

We assume that there are $K$ latent relations. Each relation $k$ 
is assigned to a mention pair $(m_i, m_j)$ with a non-negative weight (`confidence') $\alpha_{ijk}$. 
The pairwise score $(m_i, m_j)$ is computed as a weighted sum of 
relation-specific pairwise scores 
(see Figure~\ref{fig:multi-relation}, top):
\begin{equation*}
	\Phi(e_i, e_j, D) = \sum_{k=1}^K \alpha_{ijk} \Phi_k(e_i, e_j, D)
\end{equation*}
$\Phi_k(e_i, e_j, D)$ can be any pairwise score function, but here
we adopt the one from  Equation~\ref{equ:GH pairwise score}. Namely, we represent each relation $k$ by a 
diagonal matrix $\mathbf{R}_k \in \mathbb{R}^{d \times d}$, and 
\begin{equation*}
	\Phi_k(e_i, e_j, D) = \mathbf{e}_i^T \mathbf{R}_k \mathbf{e}_j
\end{equation*}
The weights $\alpha_{ijk}$ are normalized scores: 
\begin{equation}
	\label{equ:alpha}
	\alpha_{ijk} = \frac{1}{Z_{ijk}} \exp\left\{\frac{f^T(m_i,c_i)\mathbf{D}_k f(m_j,c_j)}{\sqrt{d}}\right\} 
\end{equation}
where $Z_{ijk}$ is a normalization factor, $f(m_i, c_i)$ is a function mapping $(m_i, c_i)$
onto $\mathbb{R}^d$, and $\mathbf{D}_k \in \mathbb{R}^{d \times d}$ is a diagonal matrix. 

\begin{figure}[h]
	\centering
    \includegraphics[width=0.5\textwidth]{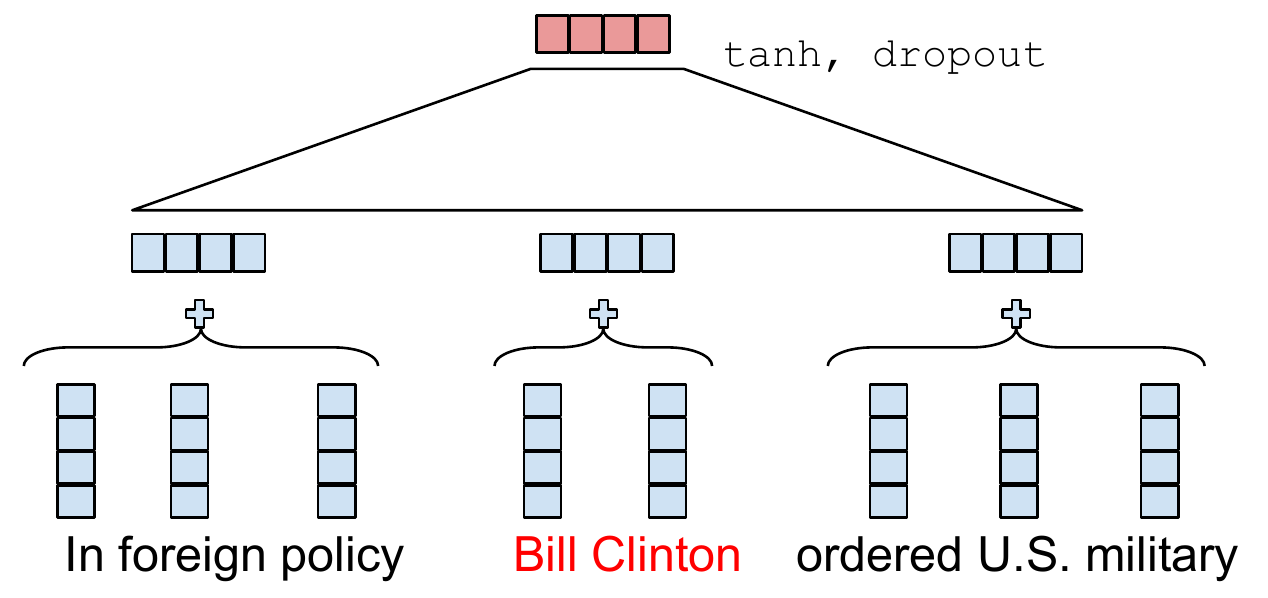}
    \caption{Function $f(m_i, c_i)$ is a single-layer neural network, 
    with $\tanh$ activation function and a layer of dropout on top.}
    \label{fig:multi-relation-f}
\end{figure}

In our experiments, we use a single-layer neural network as $f$ 
(see Figure~\ref{fig:multi-relation-f}) where 
$c_i$ is a concatenation of the average embedding of words in the left context with the average embedding of words in the right context of the mention.\footnote{We also experimented with LSTMs but we could not prevent them from severely overfitting, and the results were poor.}

As $\alpha_{ijk}$ is indexed both by mention index $j$ and relation index $k$, 
we have two choices for $Z_{ijk}$: normalization over relations 
and normalization over mentions. We consider both versions of the model.

\subsection{Rel-norm: Relation-wise normalization}
\label{subsec:k-norm}
For rel-norm, coefficients $\alpha_{ijk}$ are normalized over relations $k$, in other words,
\begin{equation*}
	Z_{ijk} = \sum_{k'=1}^K \exp\left\{\frac{f^T(m_i,c_i)\mathbf{D}_{k'} f(m_j,c_j)}{\sqrt{d}}\right\}
\end{equation*}
so that $\sum_{k=1}^K \alpha_{ijk} = 1$ (see Figure~\ref{fig:multi-relation}, middle).
We can also re-write the pairwise scores as
\begin{equation}
	\label{equ:k-normalize}
	\Phi(e_i, e_j, D) = \mathbf{e}^T_i \mathbf{R}_{ij} \mathbf{e}_j 
\end{equation}
where $\mathbf{R}_{ij} = \sum_{k=1}^K \alpha_{ijk} \mathbf{R}_k$.

Intuitively, $\alpha_{ijk}$ is the probability of assigning a $k$-th relation to a mention pair $(m_i, m_j)$. For every pair rel-norm uses these probabilities to choose one relation from the pool and relies on the corresponding relation embedding $\mathbf{R}_k$ to compute the compatibility score.

For $K=1$ rel-norm reduces (up to a scaling factor) to the bag-of-entities model
defined in Equation~\ref{equ:GH pairwise score}.

In principle, instead of relying on the linear combination of  
relation embeddings matrices $\mathbf{R}_k$, we could directly predict 
a context-specific relation embedding $\mathbf{R}_{ij} = diag\{ g(m_i, c_i, m_j, c_j) \}$
where $g$ is a neural network. However, in preliminary experiments 
we observed that this resulted in overfitting and poor performance.
Instead, we choose to use a small fixed number of relations 
as a way to constrain the model and improve generalization.

\subsection{Ment-norm: Mention-wise normalization}
\label{subsec:j-norm}
We can also normalize $\alpha_{ijk}$ over $j$:
\begin{equation*}
	Z_{ijk} = \sum_{\substack{j'=1\\j'\neq i}}^n \exp\left\{\frac{f^T(m_i,c_i)\mathbf{D}_{k} f(m_{j'},c_{j'})}{\sqrt{d}}\right\}
\end{equation*}
This implies that $\sum_{j=1,j \neq i}^n \alpha_{ijk} = 1$ (see Figure~\ref{fig:multi-relation}, bottom). 
If we rewrite the pairwise scores as
\begin{equation}
	\label{equ:j-normalize}
	\Phi(e_i, e_j, D) = \sum_{k=1}^K \alpha_{ijk} \mathbf{e}^T_i \mathbf{R}_{k} \mathbf{e}_j,
\end{equation}
we can see that
Equation~\ref{equ:GH pairwise score} 
is a special case of ment-norm 
when $K=1$ and $\mathbf{D}_1 = \mathbf{0}$.
In other words, \newcite{D17-1276} is our mono-relational ment-norm 
with uniform $\alpha$. 

The intuition behind ment-norm is that for each relation $k$ and mention $m_i$, 
we are looking for mentions related to $m_i$ 
with relation $k$. For each pair of $m_i$ and $m_j$ we can distinguish two cases: 
(i) $\alpha_{ijk}$ is small for all $k$: $m_i$ and $m_j$ are not related under any relation, 
(ii) $\alpha_{ijk}$ is large for one or more $k$: there are one or more relations which are predicted for $m_i$ and $m_j$.

In principle, rel-norm can also  indirectly handle both these cases. For example, it 
can master (i) by dedicating a distinct `none' relation to represent lack of relation between the two mentions (with the corresponding matrix $\mathbf{R}_k$ set to $\mathbf{0}$). Though it cannot assign large weights (i.e., close to $1$) to multiple relations (as needed for (ii)), it can in principle use the `none' relation to vary the probability mass assigned to the rest of relations across mention pairs, thus achieving the same effect (up to a multiplicative factor).  Nevertheless, in contrast to  ment-norm, we do not observe this behavior for rel-norm in our experiments: the inductive basis seems to disfavor such configurations. 


Ment-norm is in line with the current trend of using the attention mechanism in deep learning~\cite{bahdanau2014neural}, and especially related to multi-head attention of  \newcite{NIPS2017_7181}. For each mention $m_i$ and for each 
$k$, we can interpret $\alpha_{ijk}$ as the probability of choosing 
a mention $m_j$ among the set of mentions in the document. 
Because here we have $K$ relations, each mention $m_i$ will 
have maximally $K$ mentions 
(i.e. heads in terminology of \newcite{NIPS2017_7181}) to focus on. Note though that they use multi-head attention for choosing input features in each layer, whereas we rely on this mechanism to compute pairwise scoring functions for the structured output (i.e. to compute potential functions in the corresponding undirected graphical model, see Section~\ref{sect:impl}). 

\subsubsection*{Mention padding}
A potentially serious drawback of ment-norm is that the model uses all $K$ 
relations even in cases where some relations are inapplicable. 
For example, consider applying relation 
\emph{coreference} to mention ``West Germany'' in Figure~\ref{fig:multi-rel example}. The mention is non-anaphoric:
there are no mentions co-referent with it. Still the ment-norm model has to 
distribute the weight across the mentions. This problem occurs because 
of the normalization $\sum_{j=1,j \neq i}^n \alpha_{ijk} = 1$. Note that this  issue does not affect standard applications of attention: normally the attention-weighted signal is input to another transformation (e.g.,  a flexible neural model) which can then disregard this signal when it is useless. This is not possible within our model, as it simply uses $\alpha_{ijk}$ to weight the bilinear terms without any extra transformation.

Luckily, there is an easy way to circumvent this problem. We  add to each document 
a padding mention $m_{pad}$ linked to a padding entity $e_{pad}$.
In this way, the model can use the padding mention to damp the probability mass  
that the other mentions receive. 
This method is similar to the way some mention-ranking coreference models  
deal 
with non-anaphoric mentions (e.g. \newcite{P15-1137}).

\subsection{Implementation}
\label{sect:impl}
Following \newcite{D17-1276} we use Equation~\ref{equ:global model} 
to define a conditional random field (CRF).  We use the local score function identical to theirs and the pairwise scores are defined as explained above: 
\begin{equation*}
q(E|D) \propto \exp \left\{ \sum_{i=1}^n \Psi(e_i, c_i) + 
                     \sum_{i \neq j} \Phi(e_i, e_j, D) \right\}
\end{equation*}
We also use max-product loopy belief propagation (LBP) to estimate the max-marginal
probability
\begin{equation*}
\hat{q}_i(e_i|D) \approx \max_{\substack{e_1,...,e_{i-1}\\e_{i+1},...,e_n}} q(E|D)  
\end{equation*} 
for each mention $m_i$. 
The final score function for $m_i$ is given by:
\begin{equation*}
	\rho_i(e) = g(\hat{q}_i(e|D), \hat{p}(e|m_i))
\end{equation*}
where $g$ is a two-layer neural network and $\hat{p}(e|m_i)$ is 
the probability of selecting $e$ conditioned only on $m_i$. This probability is computed by mixing mention-entity hyperlink count statistics 
from Wikipedia, a large Web corpus and YAGO.\footnote{See \newcite[Section 6]{D17-1276}.}

We minimize the following ranking loss: 
\begin{align}
L(\theta) &= \sum_{D \in \mathcal{D}} \sum_{m_i \in D} \sum_{e \in C_i} h(m_i, e) \label{equ:loss} \\
h(m_i, e) &= \max\big(0, \gamma - \rho_i(e_i^*) + \rho_i(e)\big) \nonumber
\end{align}
where $\theta$ are the model parameters, $\mathcal{D}$ is a training dataset, 
and $e_i^*$ is the ground-truth entity. 
Adam \cite{kingma2014adam} is used as an optimizer.
 
For ment-norm, the padding mention is treated like any other mentions. 
We add $\mathbf{f}_{pad} = f(m_{pad}, c_{pad})$ and 
$\mathbf{e}_{pad} \in \mathbb{R}^d$, 
an embedding of $e_{pad}$,
to the model parameter list, and tune them while training the model. 

In order to encourage the models to explore different relations, 
we add  the following  regularization term to the loss
function in Equation~\ref{equ:loss}: 
\begin{equation*}
	\lambda_1 \sum_{i,j} \text{dist}(\mathbf{R}_i, \mathbf{R}_j)
    + \lambda_2 \sum_{i,j} \text{dist}(\mathbf{D}_i, \mathbf{D}_j)
\end{equation*}
where $\lambda_1, \lambda_2$ are set to $-10^{-7}$ in our experiments, $\text{dist}(\mathbf{x}, \mathbf{y})$ can be any distance metric. We use:
\begin{equation*}
	\text{dist}(\mathbf{x}, \mathbf{y}) = \left \Vert  \frac{\mathbf{x}}{\|\mathbf{x}\|_2} -
    							\frac{\mathbf{y}}{\|\mathbf{y}\|}_2 \right \Vert _2
\end{equation*}

Using this regularization to favor diversity is important as otherwise relations tend to collapse: their relation embeddings $\mathbf{R}_k$ end up being very similar to each other.

\section{Experiments}

We evaluated four models: 
(i) {\it rel-norm} proposed in Section~\ref{subsec:k-norm};
(ii) {\it ment-norm} proposed in Section~\ref{subsec:j-norm};
(iii) {\it ment-norm ($K=1$)}: the mono-relational version of ment-norm; and
(iv) {\it ment-norm (no pad)}: the ment-norm without using  mention padding.
Recall also that our mono-relational (i.e. $K=1$) rel-norm 
is equivalent to the relation-agnostic baseline of \newcite{D17-1276}.

We implemented our models in PyTorch
and run experiments on a Titan X GPU. The source code and 
trained models will be publicly available at 
\url{https://github.com/lephong/mulrel-nel}.

\subsection{Setup}
We set up our experiments similarly to those of \newcite{D17-1276}, 
run each model 5 times, and report average and 95\%
confidence interval of the standard micro F1 score 
(aggregates over all mentions).

\subsubsection*{Datasets}
For \emph{in-domain} scenario, we used AIDA-CoNLL dataset\footnote{TAC KBP 
datasets are no longer available.} \cite{D11-1072}. This dataset contains 
AIDA-train for training, AIDA-A for dev, and AIDA-B for testing, 
having respectively 946, 216, and 231 documents. 
For \emph{out-domain} scenario, we  evaluated the models trained on AIDA-train, on five popular test sets: 
MSNBC, AQUAINT, ACE2004, which were cleaned and updated by 
\newcite{guorobust}; 
WNED-CWEB (CWEB), WNED-WIKI (WIKI), which were automatically extracted 
from ClueWeb and Wikipedia \cite{guorobust,gabrilovich2013facc1}.
The first three are small with 20, 50, and 36 documents whereas 
the last two are much larger with 320 documents each. 
Following previous works \cite{K16-1025, D17-1276}, we considered 
only mentions that have entities in the KB (i.e., Wikipedia). 



\subsubsection*{Candidate selection}
For each mention $m_i$, we selected 30 top candidates 
using $\hat{p}(e | m_i)$. We then kept 4 candidates with the highest 
$\hat{p}(e | m_i)$ and 3 candidates with the highest scores
$\mathbf{e}^T \left( \sum_{w \in d_i} \mathbf{w} \right)$,
where $\mathbf{e}, \mathbf{w} \in \mathbb{R}^d$ are entity and word 
embeddings, $d_i$ is the 50-word window context around $m_i$.

\subsubsection*{Hyper-parameter setting}

We set $d=300$ and used GloVe \cite{D14-1162} word embeddings trained on 840B 
tokens
for computing $f$ in Equation~\ref{equ:alpha}, 
and entity embeddings from \newcite{D17-1276}.\footnote{\url{https://github.com/dalab/deep-ed}}
We use the following parameter values: $\gamma=0.01$ (see  Equation~\ref{equ:loss}), 
the number of LBP loops is 10, 
the dropout rate for $f$ was set to 0.3, 
the window size of local contexts $c_i$ (for the pairwise score functions) is 6.
For rel-norm, we initialized $\text{diag}(\mathbf{R}_k)$ and 
$\text{diag}(\mathbf{D}_k)$ by 
sampling from $\mathcal{N}(0, 0.1)$ for all $k$. 
For ment-norm, we did the same except that
$\text{diag}(\mathbf{R}_1)$ was sampled from $\mathcal{N}(1, 0.1)$. 

To select the best number of relations $K$, we considered all values of 
$K \le 7$ ($K > 7$ would not fit in our GPU memory, as some of the 
documents are large). We selected the best ones based on the development scores: 
6 for rel-norm, 3 for ment-norm, and 3 for ment-norm (no pad).

When training the models, we applied early stopping. For rel-norm, 
when the model reached $91\%$ F1 on the dev set, \footnote{We chose the highest F1 
that rel-norm always achieved without the learning rate reduction.} 
we reduced the learning rate 
from $10^{-4}$ to $10^{-5}$. We then stopped the training when F1 was not 
improved after 20 epochs. We did the same for ment-norm except that 
the learning rate was changed at $91.5\%$ F1.

Note that all the hyper-parameters except $K$ and the turning point
for early stopping were set to the values used by \newcite{D17-1276}. 
Systematic tuning is expensive though may have further increased 
the result of our models.

\subsection{Results}

\begin{table}[ht]
	\small
    \centering
    \begin{tabular}{c|c}
    	Methods & Aida-B \\
        \hline
        \newcite{Q15-1011} & 88.7 \\
        \newcite{guorobust} & 89.0 \\
        \newcite{P16-1059} & 91.0 \\
        \newcite{K16-1025} & 91.5 \\
        \newcite{D17-1276} & $92.22 \pm 0.14$ \\
        \hline
        rel-norm & $92.41 \pm 0.19$ \\
        ment-norm & $\mathbf{93.07} \pm 0.27$ \\
        ment-norm ($K=1$) & $92.89 \pm 0.21$ \\
        ment-norm (no pad) & $92.37 \pm 0.26$ 
    \end{tabular}
    \caption{F1 scores on AIDA-B (test set).}
    \label{tab:in-domain test}
\end{table}

\begin{table*}[ht]
	\small
    \centering
    \begin{tabular}{c|c|c|c|c|c||c}
    	Methods & MSNBC & AQUAINT & ACE2004 & CWEB & WIKI & Avg \\
        \hline
        \newcite{milne2008learning} & 78 & 85 & 81 & 64.1 & 81.7 & 77.96 \\
        \newcite{D11-1072} & 79 & 56 & 80 & 58.6 & 63 & 67.32 \\
        \newcite{P11-1138} & 75 & 83 & 82 & 56.2 & 67.2 & 72.68 \\
        \newcite{cheng-roth:2013:EMNLP} & 90 & \textbf{90} & 86 & 67.5 & 73.4 & 81.38 \\
        \newcite{guorobust} & 92 & 87 & 88 & 77 & \textbf{84.5} & \textbf{85.7} \\
        \newcite{D17-1276} & 93.7 $\pm$ 0.1 & 88.5 $\pm$ 0.4 & 88.5 $\pm$ 0.3 & \textbf{77.9} $\pm$ 0.1 & 77.5 $\pm$ 0.1 & 85.22 \\
        \hline
        rel-norm& 92.2 $\pm$ 0.3 & 86.7 $\pm$ 0.7 & 87.9 $\pm$ 0.3 & 75.2 $\pm$ 0.5 & 76.4 $\pm$ 0.3 & 83.67 \\
        ment-norm & \textbf{93.9} $\pm$ 0.2 & 88.3 $\pm$ 0.6 & \textbf{89.9} $\pm$ 0.8 & 77.5 $\pm$ 0.1 & \underline{78.0} $\pm$ 0.1 & \underline{85.51} \\
        ment-norm ($K=1$) & 93.2 $\pm$ 0.3 & 88.4 $\pm$ 0.4 & 88.9 $\pm$ 1.0 & 77.0 $\pm$ 0.2 & 77.2 $\pm$ 0.1 & 84.94  \\
        ment-norm (no pad) & 93.6 $\pm$ 0.3 & 87.8 $\pm$ 0.5 & \textbf{90.0} $\pm$ 0.3 & 77.0 $\pm$ 0.2 & 77.3 $\pm$ 0.3 & 85.13 
    \end{tabular}
    \caption{F1 scores on five out-domain test sets. Underlined scores show 
     cases where the corresponding model outperforms the baseline.}
    \label{tab:out-domain test}
\end{table*}

Table~\ref{tab:in-domain test} shows micro F1 scores on AIDA-B
of the SOTA methods and ours, which all use 
Wikipedia and YAGO mention-entity index. 
To our knowledge, ours are the only (unsupervisedly) inducing and employing 
more than one relations on this dataset. 
The others use only one relation, coreference,
which is given by simple heuristics or supervised third-party resolvers.
All four our models outperform any previous method, with ment-norm 
achieving the best results, 0.85\% higher than 
that of \newcite{D17-1276}.

Table~\ref{tab:out-domain test} shows micro F1 scores on 5 
out-domain test sets. Besides ours, only \newcite{cheng-roth:2013:EMNLP}
employs several mention relations. 
Ment-norm achieves the highest F1 scores on 
MSNBC and ACE2004. On average, ment-norm's F1 score is 0.3\% higher 
than that of \newcite{D17-1276}, but 0.2\% lower than 
\newcite{guorobust}'s. It is worth noting that \newcite{guorobust}
performs exceptionally well on WIKI, but substantially worse 
than ment-norm on all other datasets.
Our other three models, however, 
have lower average F1 scores compared to the best previous model.  

The experimental results show that ment-norm outperforms 
rel-norm, and that mention padding plays an important role.

\subsection{Analysis}

\subsubsection*{Mono-relational v.s. multi-relational}
For rel-norm, the mono-relational version (i.e., \newcite{D17-1276}) is outperformed by 
the multi-relational one on AIDA-CoNLL, but performs significantly 
better on all five out-domain
datasets. This implies that multi-relational rel-norm does  not 
generalize well across domains. 

For ment-norm, the mono-relational version performs worse than 
the multi-relational one on all test sets except AQUAINT. 
We speculate that this is due to multi-relational ment-norm being less sensitive to prediction errors. Since it can rely on multiple factors more easily, a single mistake in assignment is unlikely to have large influence on its predictions.

\subsubsection*{Oracle}
\begin{figure}[ht]
\centering
\resizebox {0.45\textwidth}{0.3\textwidth}{
\begin{tikzpicture}
\pgfplotsset{every tick label/.append style={font=\small}}
\begin{axis}[
    ybar,
    ymajorgrids,
    enlargelimits=0.15,
    legend style={at={(0,1)},anchor=north west},
    symbolic x coords={G\&H,rel-norm,ment-norm (K=1),ment-norm},
    xticklabel style={text width=2cm,align=center},
    ]
\addplot coordinates {(G\&H,92.22) (rel-norm,92.41) (ment-norm (K=1),92.89) (ment-norm,93.07)};
\addplot coordinates {(G\&H,93.19) (rel-norm,94.15) (ment-norm (K=1),93.67) (ment-norm,94.34)};
\legend{LBP,oracle}
\end{axis}
\end{tikzpicture}
}
\caption{F1 on AIDA-B when using LBP 
and the oracle. G\&H is \newcite{D17-1276}.}
\label{fig:oracle}
\end{figure}
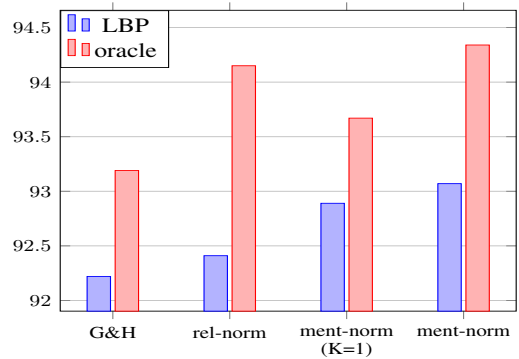

In order to examine learned relations in a more transparant setting, we consider 
an idealistic scenario where imperfection of LBP, as well as mistakes 
in predicting other entities,
are taken out of the equation
 using an oracle. This oracle, when we make a prediction 
for mention $m_i$, will tell us the correct entity $e_j^*$ 
for every other mentions $m_j, j\neq i$.
We also used AIDA-A (development set) for selecting the numbers of relations for 
rel-norm and ment-norm. They are set to 6 and 3, respectively. 
Figure~\ref{fig:oracle} shows the micro F1 scores. 

Surprisingly, the performance of oracle rel-norm is 
close to that of oracle ment-norm, although without using the oracle 
the difference was substantial. 
This suggests that rel-norm is more sensitive 
to prediction errors than ment-norm.
\newcite{D17-1276}, even with the help of the oracle, can only 
perform slightly better than LBP (i.e. non-oracle) ment-norm. 
This suggests that its global coherence scoring component is indeed too simplistic.
Also note that both multi-relational oracle models 
substantially outperform the two mono-relational oracle models. 
This shows the benefit of using more than one relations, 
and the potential of achieving higher accuracy 
with more accurate inference methods.

\subsection*{Relations}

\begin{figure*}[h]
\centering
\small
\setlength{\tabcolsep}{2pt}
\begin{tabular}{lll|l|lll}
 \multicolumn{3}{c}{rel-norm} & on Friday , \textbf{Liege} police said in & \multicolumn{3}{c}{ment-norm} \\ 
 \hline
\mybar{0.92}{0.03}{0.05} & (1) missing teenagers in \textbf{Belgium} . & \mybar{0.03}{0.15}{0.14} \\
\mybar{0.32}{0.04}{0.64} & (2) UNK \textbf{BRUSSELS} UNK & \mybar{0.03}{0.16}{0.15} \\
\mybar{0.76}{0.13}{0.11} & (3) UNK \textbf{Belgian} police said on & \mybar{0.04}{0.12}{0.12} \\
\mybar{0.94}{0.02}{0.04} & (4) , " a \textbf{Liege} police official told & \mybar{0.27}{0.05}{0.06} \\
\mybar{0.8}{0.05}{0.15} & (5) police official told \textbf{Reuters} . & \mybar{0.04}{0.12}{0.11} \\
\mybar{0.95}{0.02}{0.03} & (6) eastern town of \textbf{Liege} on Thursday , & \mybar{0.25}{0.05}{0.06} \\
\mybar{0.7}{0.08}{0.22} & (7) home village of \textbf{UNK} . & \mybar{0.1}{0.08}{0.08} \\
\mybar{0.39}{0.12}{0.49} & (8) link with the \textbf{Marc Dutroux} case , the & \mybar{0.11}{0.08}{0.08} \\
\mybar{0.82}{0.11}{0.07} & (9) which has rocked \textbf{Belgium} in the past & \mybar{0.04}{0.12}{0.12} \\
\end{tabular}
\raisebox{-.5\height}{\includegraphics[width=0.4\textwidth]{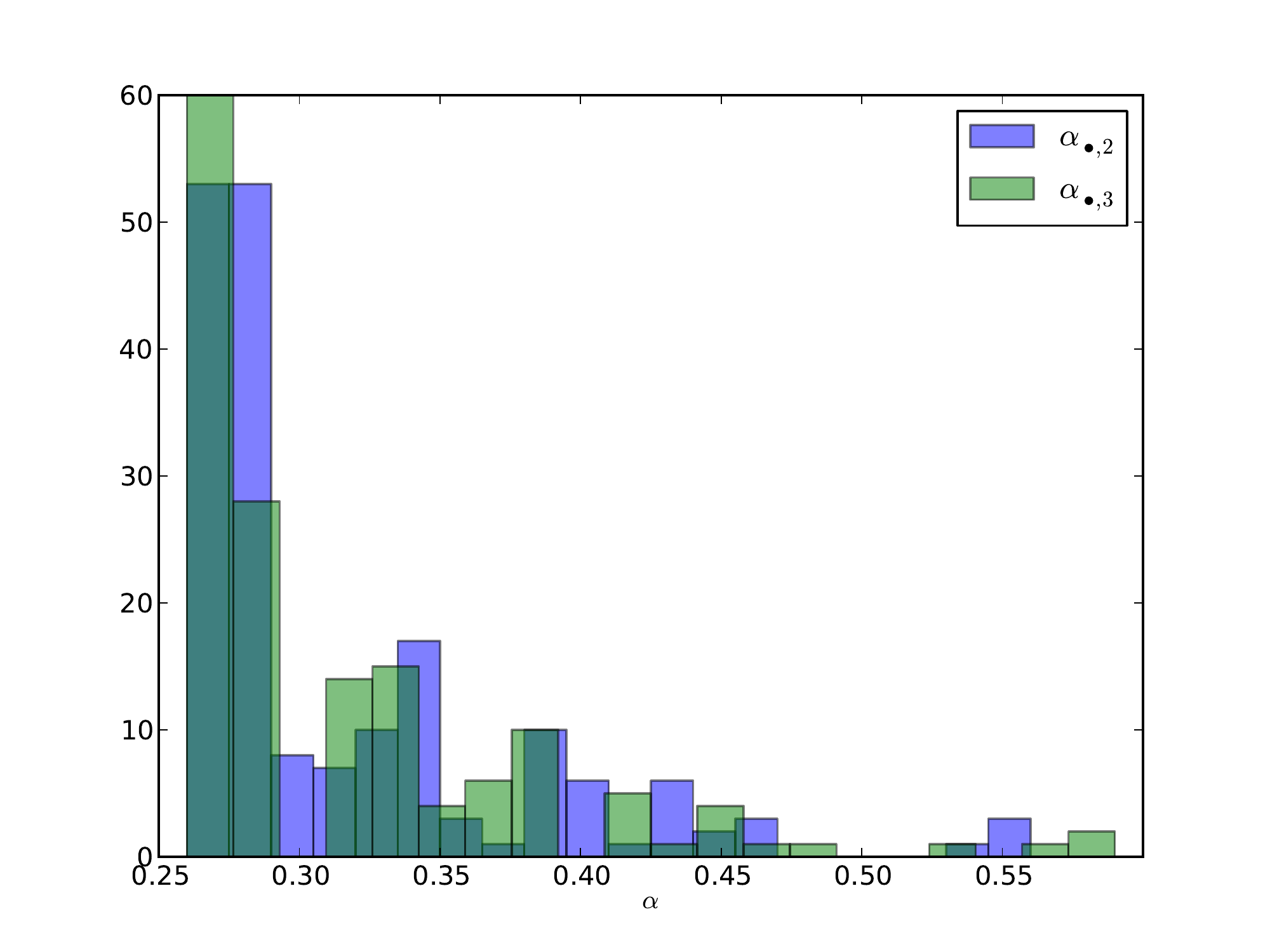}}
\caption{(Left) Examples of $\alpha$. The first and third columns 
show $\alpha_{ijk}$ for oracle rel-norm and oracle ment-norm, respectively.
(Right) Histograms of $\alpha_{\bullet k}$ for $k=2,3$, 
corresponding to the second and third relations from oracle ment-norm. 
Only $\alpha > 0.25$ (i.e. high attentions) are shown.}
\label{tab:demo alpha}
\end{figure*}

In this section we qualitatively examine relations that the models learned by looking at the probabilities $\alpha_{ijk}$.
See Figure \ref{tab:demo alpha} for an example.  In that example we focus on mention ``Liege'' in the sentence at the top and study which mentions are related to it under two versions of our model: rel-norm (leftmost column) and ment-norm (rightmost column). 

For rel-norm it is difficult to interpret the meaning of the relations.
It seems that the first relation 
dominates the other two, with very high weights for most of the mentions. 
Nevertheless, the fact that rel-norm outperforms the baseline 
suggests that those learned relations encode some useful information.

For ment-norm, the first relation is 
similar to coreference: the relation prefers those mentions that potentially refer to the same entity (and/or have semantically similar mentions): see
Figure \ref{tab:demo alpha} (left, third column).
The second and third relations behave differently from the first relation
as they prefer mentions having more distant meanings
and are complementary to the first relation. They 
assign large weights to (1) ``Belgium'' and (2) ``Brussels'' 
but small weights to (4) and (6) ``Liege''. The two relations look similar in this example, however they are not identical in general. See a histogram of bucketed values of their weights in
Figure~\ref{tab:demo alpha} (right): their $\alpha$ have 
quite different distributions. 

\subsubsection*{Complexity}
The complexity of rel-norm and ment-norm is linear in
$K$, so in principle our models should be considerably more expensive than \newcite{D17-1276}. 
However, our models
converge much faster than their relation-agnostic model: on average 
ours needs 120 epochs, compared to theirs 1250 epochs. We believe that the structural bias helps the model to capture necessary regularities more easily.
In terms of wall-clock time, our model requires just under 1.5 hours to train, 
that is ten times faster than the relation agnostic model~\cite{D17-1276}.
In addition, the difference in testing time is negligible 
when using a GPU.

\section{Conclusion and Future work}
We have shown the benefits of using relations in NEL. Our models consider 
relations as latent variables, thus do not require any extra 
supervision. Representation learning was used to learn 
relation embeddings, eliminating the need for extensive feature engineering. 
The experimental results show that 
our best model achieves the best reported F1 on AIDA-CoNLL
with an improvement of 0.85\% F1 over the best previous results. 

Conceptually, modeling multiple relations is substantially different 
from simply modeling coherence (as in \newcite{D17-1276}). 
In this way we also hope it will lead to interesting follow-up work, 
as individual relations can be informed by injecting prior knowledge 
(e.g., by training jointly with relation extraction models).

In future work, we would like to use syntactic and discourse 
structures (e.g., syntactic dependency paths between mentions)
to encourage the models to discover a richer set of  relations.
We also would like to combine ment-norm and rel-norm. Besides, 
we would like to examine whether the induced latent relations 
could be helpful for relation extract.

\section*{Acknowledgments}
We would like to thank anonymous reviewers for their 
suggestions and comments. The project was supported by the
European Research Council (ERC StG BroadSem
678254), the Dutch National Science Foundation
(NWO VIDI 639.022.518), and an Amazon Web Services
(AWS) grant.

\bibliography{ref}
\bibliographystyle{acl_natbib}

\appendix


\end{document}